\documentclass{article}

\PassOptionsToPackage{numbers, compress}{natbib}

\usepackage[final]{nips_2018}

\usepackage[utf8]{inputenc} 
\usepackage[T1]{fontenc}    
\usepackage{hyperref}       
\usepackage{url}            
\usepackage{booktabs}       
\usepackage{amsfonts}       
\usepackage{nicefrac}       
\usepackage{microtype}      

\usepackage{times}
\usepackage{epsfig}
\usepackage{graphicx}
\usepackage{amsmath}
\usepackage{amssymb}
\usepackage{subcaption}
\usepackage[export]{adjustbox}

\title{An Interpretable Model for Scene Graph Generation}

\author{
Ji Zhang\textsuperscript{1,2},
Kevin Shih\textsuperscript{1},
Andrew Tao\textsuperscript{1},
Bryan Catanzaro\textsuperscript{1},
Ahmed Elgammal\textsuperscript{2} \\
\\
\textsuperscript{1}Nvidia Research \\
\textsuperscript{2}Department of Computer Science, Rutgers University}

\begin{document}

\maketitle

\begin{abstract}
  We propose an efficient and interpretable scene graph generator. We consider three types of features: visual, spatial and semantic, and we use a late fusion strategy such that each feature's contribution can be explicitly investigated. We study the key factors about these features that have the most impact on the performance, and also visualize the learned visual features for relationships and investigate the efficacy of our model. We won the champion of the OpenImages Visual Relationship Detection Challenge on Kaggle, where we outperform the 2nd place by 5\% (20\% relatively). We believe an accurate scene graph generator is a fundamental stepping stone for higher-level vision-language tasks such as image captioning and visual QA, since it provides a semantic, structured comprehension of an image that is beyond pixels and objects.
\end{abstract}

\section{Introduction}

Scene graph generation is a fundamental task that bridges low-level vision such as scene parsing and object detection and high-level vision-language problems such as image captioning \cite{Lu2018Neural, yao2018exploring} and visual QA \cite{antol2015vqa, johnson2017clevr}. It produces a structured semantic understanding of an image from individual objects, and provides rich information for those high-level tasks. Current state-of-the-art methods \cite{lu2016visual, yu17iccv, Zhuang_2017_ICCV, plummerPLCLC2017, dai2017detecting, zhang2017visual, Yang_2018_ECCV, LiCVPR2017, xu2017scenegraph, zellers2018neural, Yin_2018_ECCV, zhang2017relationship,zhang2018large} use three types of features to represent relationships: 1) \textit{visual features}: the CNN features of the two objects or their combination; 2) \textit{spatial features}: coordinates of the two objects which encodes their spatial layouts; 3) \textit{semantic features}: class labels of the two objects which provide a strong prior of the predicate. Most of them, if not all, combine the three features in an early stage to learn a compositional feature for relationship prediction. The contribution of each feature is thus implicit and probably not optimized. In this paper we propose a structure that instead explicitly builds three branches for the three features, each contributing to the output in an interpretable way, and we fuse their outputs in the final stage to get optimized predictions.

Our contributions are: 1) we propose a new model that efficiently combines three features and show explicitly what each feature contributes to the final prediction and how much the contribution is. 2) we demonstrate the efficacy of our model on three datasets: OpenImages (OI) \cite{openimages}, Visual Genome (VG) \cite{krishnavisualgenome} and Visual Relationship Detection (VRD) \cite{lu2016visual}. We won the 1st place in the OpenImages Challenge, and we outperform state-of-the-art methods on VG and VRD by significant margins.

\section{Model Description}

The task of visual relationship detection can be defined as a mapping $f$ from image $I$ to 3 labels and 2 boxes $l_S,l_P,l_O,b_S,b_O$
\begin{equation}
I \xrightarrow{f} l_S,l_P,l_O,b_S,b_O
\end{equation}
where $l,b$ stand for labels and boxes, $S,P,O$ stand for subject, predicate, object. We decompose $f$ into object detector $f_{det}$ and relationship classifier $f_{rel}$:
\begin{equation}
I \xrightarrow{f_{det}} l_S,l_O,b_S,b_O,v_S,v_O \xrightarrow{f_{rel}} l_P
\end{equation}
The decomposition means that we can run an object detector on the input image to obtain labels, boxes and visual features for subject and object, then use these as input features to the relationship classifier which only needs to output a label. There are two obvious advantages in this model: 1) learning complexity is dramatically reduced, since we can simply use an off-the-shelf object detector as $f_{det}$ without the need for re-training, hence the learn-able weights exist only in the small subnet $f_{rel}$; 2) We have much richer features for relationships, i.e., $l_S,l_O,b_S,b_O,v_S,v_O$ for $f_{rel}$, instead of only the image $I$ for $f$.

We further assume that the semantic feature $l_S,l_O$, spatial feature $b_S,b_O$ and visual feature $v_S,v_O$ are independent from each other. So we can build 3 separate branches of sub-networks for them. This is the basic work flow of our model.

\begin{figure*}
  \centering
  \includegraphics[width=\textwidth]{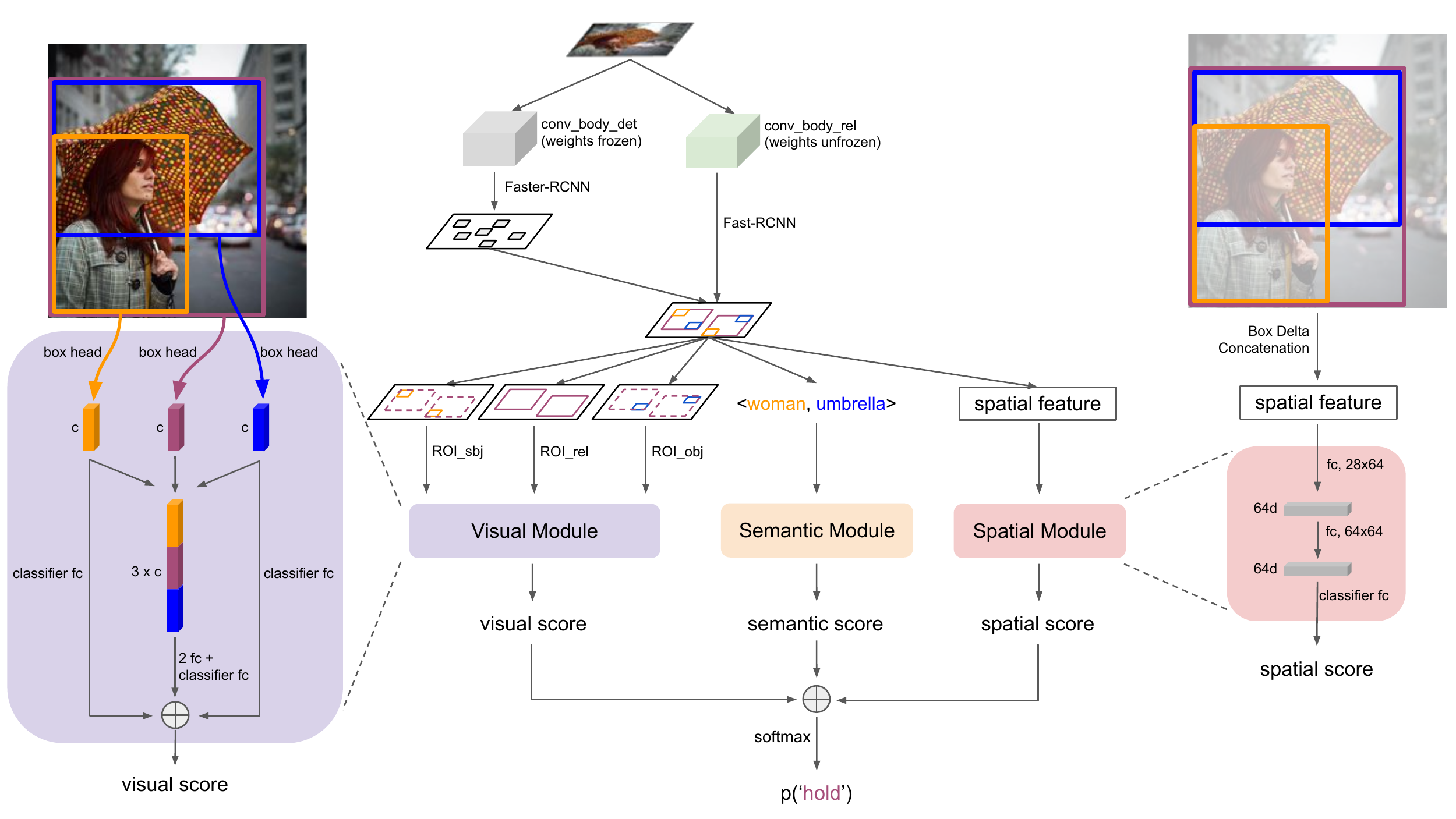}
  \setlength{\abovecaptionskip}{-5pt}
  \setlength\belowcaptionskip{-2ex}
  \caption{Model Architecture}
  \label{fig:architecture}
\end{figure*}

Figure \ref{fig:architecture} shows our model in details. The network takes an input image and outputs the 6 aforementioned features, then each branch uses its corresponding feature to produce a confidence score for predicates, then all scores are added up and normalized by softmax. We now introduce each module's design and their motivation.

\subsection{Relationship Proposal}

A relationship proposal is defined as a pair of objects that is very likely related\cite{zhang2017relationship}. In our model we first detect all meaningful objects by running an object detector, then we simply consider each pair of objects is a relationship proposal. The following modules learn to classify each pair as either ``no relationship'' or one of the $9$ predicates, not including the ``is'' relationship.

\subsection{Semantic Module}

Zeller, et al.\cite{zellers2018neural} introduced a frequency baseline that performs reasonably well on Visual Genome dataset\cite{xu2017scene} by counting frequencies of predicates given subject and object. Its motivation is that in general cases, the types of relationships between two objects are usually limited, e.g., given the subject being person and object being horse, their relationship is highly likely to be “ride”, “walk”, “feed”, but less likely to be “stand on”, “carry”, “wear”, etc. In short, the $\langle subject,predicate,object\rangle$ composition is usually biased. Furthermore, any specific relationship detection dataset can only contain a limited number of them, making the bias even stronger. This is a factor that we find every useful to leverage.

We improved this baseline by removing the background class of subject and object. Specifically, for each training image we count the occurrence of $l_P$ given $l_S,l_O$ in the ground truth annotations, and we end up with an empirical distribution $p(P|S,O)$ for the whole training set. We do this under the assumption that the test set is also drawn from the same distribution. We then build the remaining modules to learn a complementary residual on top of the output of this baseline.

\subsection{Spatial Module}

In the challenge dataset, the three predicates ``on'', ``under'', ``inside\_of'' indicate purely spatial relationships i.e., the relative locations of subject and object are sufficient to tell the relationship. A common solution, as applied in Faster-RCNN\cite{ren2015faster}, is to learn a mapping from visual features to location offsets. However, the learning becomes significantly hard when the distance of two objects are very far\cite{gkioxari2017interactnet}, which is often the case for relationships. We capture spatial information by encoding the box coordinates of subjects and objects using box delta\cite{ren2015faster} and normalized coordinates:
\begin{equation}
\langle\Delta(b_S,b_O),\Delta(b_S,b_P),\Delta(b_P,b_O),c(b_S),c(b_O)\rangle
\end{equation}
where $\Delta(b_1,b_2)$ are box delta of two boxes $b_1,b_2$, and $c(b)$ are normalized coordinates of box $b$, which are defined as:
\begin{equation}
\Delta(b_1,b_2)=\langle\frac{x_1-x_2}{w_2},\frac{y_1-y_2}{h_2},\log\frac{w_1}{w_2},\log{h_1}{h_2}\rangle
\end{equation}
\begin{equation}
c(b)=\langle\frac{x_{min}}{w},\frac{y_{min}}{h},\frac{x_{max}}{w},\frac{y_{max}}{h},\frac{a_{box}}{a_{img}}\rangle
\end{equation}
where $b_1=(x_1,y_1,w_1,h_1)$ and $b_2=(x_2,y_2,w_2,h_2)$, $w,h$ are width and height of the image, $a_{box}$ and $a_{img}$ are areas of the box and image.

\subsection{Visual Module}

Visual Module is useful mainly for three reasons: 1) it accounts for all other types of relationships that spatial features can hardly predict, e.g., interactions such as ``man play guitar'' and ``woman wear handbag''; 2) it solves relationship reference problems\cite{krishna2018referring}, i.e., when there are multiple subjects or objects that belong to a same category, we need to know which subject is related to which object; 3) for some specific interactions, e.g., ``throw'', ``eat'' ``ride'', the visual appearance of the subject or object alone is very informative about the predicate. With these motivations, we feed subject, predicate, object ROIs into the backbone and get the feature vectors from its last fc layer as our visual features, then we concatenate these three features and feed them into 2 additional randomly initialized fc layers followed by an extra fc layer to get a logit, i.e., unnormalized score. We also add one fc layer on top of the subject feature and another fc layer on top of the object feature to get two scores. These two scores are the predictions made solely by the subject/object feature according to the third reason mentioned above.

\subsection{The ``is'' Relationship}

In the OpenImages challenge, ``$\langle object \rangle$ is $\langle attribute \rangle$'' is also considered as relationships, where there is only one object involved. We achieve this sub-task by using a completely separate, single-branch, Fast-RCNN based model. We use the same object detector to get proposals for this model, then for each proposal the model produces a probability distribution over all attributes with the Fast-RCNN pipeline.

\section{Experiments}
We present quantitative and qualitative results on OpenImages (OI). We show ablation study on each component of our model. We also show results on Visual Genome (VG) and Visual Relationship Detection (VRD) datasets compared with strong previous methods.

\begin{table}[t]
\centering
\begin{minipage}{0.35\linewidth}
\begin{center}
\resizebox{\columnwidth}{!}{
\begin{tabular}{c c c}
\hline
Team ID & Public \\
\hline
VRD\_NN (8th) & 0.20643 \\
anokas (7th) & 0.21573 \\
MIL (6th) & 0.21774 \\
tito (5th) & 0.25571 \\
toshif (4th) & 0.25621 \\
Kyle (3rd) & 0.28043 \\
radek (2nd) & 0.28886 \\
Seiji (Ours) & \textbf{0.33213} \\
\hline
\end{tabular}
}
\label{tab:oi_public}
\end{center}
\end{minipage}
\hspace{0.3cm}
\begin{minipage}{0.425\linewidth}
\begin{center}
\resizebox{\columnwidth}{!}{
\begin{tabular}{c c c}
\hline
Team ID & Private \\
\hline
[ods.ai] ZFTurbo (8th) & 0.17621 \\
anokas (7th) & 0.17960 \\
MIL (6th) & 0.19666 \\
radek (5th) & 0.20113 \\
toshif (4th) & 0.22832 \\
Kyle (3rd) & 0.23491 \\
tito (2nd) & 0.23709 \\
Seiji (Ours) & \textbf{0.28544} \\
\hline
\end{tabular}
}
\label{tab:oi_private}
\end{center}
\end{minipage}
\captionsetup{font=small}
\setlength\belowcaptionskip{-2ex}
\caption{\rule{0pt}{10pt} Kaggle leader boards.}
\label{tab:oi}
\end{table}

\begin{table}[h!]
\centering
\resizebox{0.8\columnwidth}{!}{
\begin{tabular}{c c c c c}
\hline
 & R@50 & mAP\_rel & mAP\_phr & score \\
\hline
 Baseline & 72.98 & 26.54 & 32.77 & 38.32 \\
 $\langle S,P,O \rangle$ & 74.13 & 32.41 & 39.55 & 43.61 \\
 $\langle S,P,O \rangle + S + O$ & \textbf{74.46} & 34.16 & 39.59 & 44.39 \\
 $\langle S,P,O \rangle + S + O + spt$ & 74.40 & \textbf{34.96} & \textbf{40.70} & \textbf{45.14} \\
\hline
\end{tabular}
}
\captionsetup{font=small}
\setlength\belowcaptionskip{-2ex}
\caption{\rule{0pt}{10pt} Ablation Study on OI.}
\label{tab:abl}
\end{table}


\begin{table}[t!]
\centering
\resizebox{\columnwidth}{!}{
\begin{tabular}{l c c c | c c c | c c c | c c | c c | c c}
\hline
& \multicolumn{9}{c}{Graph Constraint} & \multicolumn{6}{c}{No Graph Constraint} \\
& \multicolumn{3}{c}{SGDET} & \multicolumn{3}{c}{SGCLS} & \multicolumn{3}{c}{PRDCLS} & \multicolumn{2}{c}{SGDET} & \multicolumn{2}{c}{SGCLS} & \multicolumn{2}{c}{PRDCLS} \\
Recall at & 20 & 50 & 100 & 20 & 50 & 100 & 20 & 50 & 100 & 50 & 100 & 50 & 100 & 50 & 100 \\
\hline
VRD\cite{lu2016visual} & - & 0.3 & 0.5 & - & 11.8 & 14.1 & - & 27.9 & 35.0 & - & - & - & - & - & - \\
Associative Embedding\cite{newell2017pixels} & 6.5 & 8.1 & 8.2 & 18.2 & 21.8 & 22.6 & 47.9 & 54.1 & 55.4 & 9.7 & 11.3 & 26.5 & 30.0 & 68.0 & 75.2 \\
Message Passing\cite{xu2017scenegraph} & - & 3.4 & 4.2 & - & 21.7 & 24.4 & - & 44.8 & 53.0 & - & - & - & - & - & - \\
Message Passing+ & 14.6 & 20.7 & 24.5 & 31.7 & 34.6 & 35.4 & 52.7 & 59.3 & 61.3 & 22.0 & 27.4 & 43.4 & 47.2 & 75.2 & 83.6 \\
Frequency & 17.7 & 23.5 & 27.6 & 27.7 & 32.4 & 34.0 & 49.4 & 59.9 & 64.1 & 25.3 & 30.9 & 40.5 & 43.7 & 71.3 & 81.2 \\
Frequency+Overlap & 20.1 & 26.2 & 30.1 & 29.3 & 32.3 & 32.9 & 53.6 & 60.6 & 62.2 & 28.6 & 34.4 & 39.0 & 43.4 & 75.7 & 82.9 \\
MotifNet-NOCONTEXT & 21.0 & 26.2 & 29.0 & 31.9 & 34.8 & 35.5 & 57.0 & 63.7 & 65.6 & 29.8 & 34.7 & 43.4 & 46.6 & 78.8 & 85.9 \\
MotifNet-LeftRight & \textbf{21.4} & 27.2 & 30.3 & 32.9 & 35.8 & 36.5 & 58.5 & 65.2 & 67.1 & \textbf{30.5} & 35.8 & 44.5 & 47.7 & 81.1 & 88.3 \\
\textbf{Ours} & 20.8 & \textbf{28.1} & \textbf{32.5} & \textbf{36.1} & \textbf{36.7} & \textbf{36.7} & \textbf{66.7} & \textbf{68.3} & \textbf{68.3} & 30.1 & \textbf{36.4} & \textbf{48.9} & \textbf{50.8} & \textbf{93.7} & \textbf{97.7} \\
\hline
\end{tabular}
}
\setlength\belowcaptionskip{-2ex}
\captionsetup{font=small}
\caption{\rule{0pt}{10pt} Comparison with state-of-the-arts on VG.}
\label{tab:vg}
\end{table}

\begin{table}[t!]
\begin{adjustbox}{max width=1\textwidth,center}
\begin{tabular}{l c c c c | c c c c c c | c c c c c c}
\hline
& \multicolumn{2}{c}{Relationship} & \multicolumn{2}{c}{Phrase} & \multicolumn{6}{c}{Relationship Detection} & \multicolumn{6}{c}{Phrase Detection} \\
& \multicolumn{4}{c}{free k} & \multicolumn{2}{c}{k = 1} & \multicolumn{2}{c}{k = 10}  & \multicolumn{2}{c}{k = 70}  & \multicolumn{2}{c}{k = 1} & \multicolumn{2}{c}{k = 10}  & \multicolumn{2}{c}{k = 70} \\
Recall at & 50 & 100 & 50 & 100 & 50 & 100 & 50 & 100 & 50 & 100 & 50 & 100 & 50 & 100 & 50 & 100 \\
\hline
DR-Net*\cite{dai2017detecting} & 17.73 & 20.88 & 19.93 & 23.45 \rule{3pt}{0pt}& - & - & - & - & - & - & - & - & - & - & - & - \\
ViP-CNN\cite{LiCVPR2017} & 17.32 & 20.01 &  22.78 & 27.91 & 17.32 & 20.01 & - & - & - & - & 22.78 & 27.91 & - & - & - & - \\
VRL\cite{liang2017deep} & 18.19 & 20.79 & 21.37 & 22.60  & 18.19 & 20.79 & - & - & - & - & 21.37 & 22.60 & - & - & - & - \\
PPRFCN*\cite{zhang2017ppr} & 14.41 & 15.72 & 19.62 & 23.75  & - & - & - & - & - & - & - & - & - & - & - & - \\
VTransE* & 14.07 & 15.20 & 19.42 & 22.42  & - & - & - & - & - & - & - & - & - & - & - & - \\
SA-Full*\cite{Peyre17} & 15.80 & 17.10 & 17.90 & 19.50  & - & - & - & - & - & - & - & - & - & - & - & - \\
CAI*\cite{Zhuang_2017_ICCV} & 20.14 & 23.39 & 23.88 & 25.26 & - & - & - & - & - & - & - & - & - & - & - & - \\
KL distilation\cite{yu17iccv} & 22.68 & 31.89 & 26.47 & 29.76 & 19.17 & 21.34 & 22.56 & 29.89 & 22.68 & 31.89 & 23.14 & 24.03 & 26.47 & 29.76 & 26.32 & 29.43 \\
Zoom-Net\cite{Yin_2018_ECCV} & 21.37 & 27.30 & 29.05 & 37.34 & 18.92 & 21.41 & - & - & 21.37 & 27.30 & 24.82 & 28.09 & - & - & 29.05 & 37.34 \\
CAI + SCA-M\cite{Yin_2018_ECCV} & 22.34 & 28.52 & 29.64 & 38.39 & 19.54 & 22.39 & - & - & 22.34 & 28.52 & 25.21 & 28.89 & - & - & 29.64 & 38.39 \\
\textbf{Ours (ImageNet)} & 21.62 & 26.12 & 28.59 & 35.18 & 19.57 & 22.61 & 21.62 & 26.12 & 21.62 & 26.12 & 26.39 & 31.28 & 28.59 & 35.18 & 28.59 & 35.18 \\
\textbf{Ours (COCO)} & \textbf{26.67} & \textbf{32.55} & \textbf{33.29} & \textbf{41.25} & \textbf{24.30} & \textbf{27.91} & \textbf{26.67} & \textbf{32.55} & \textbf{26.67} & \textbf{32.55} & \textbf{31.09} & \textbf{36.42} & \textbf{33.29} & \textbf{41.25} & \textbf{33.29} & \textbf{41.25} \\
\hline
\end{tabular}
\end{adjustbox}
\setlength\belowcaptionskip{-2ex}
\captionsetup{font=small}
\caption{\rule{0pt}{10pt} Results on the VRD\cite{lu2016visual} dataset ($-$ means unavailable / unknown)}
\label{tab:vrd}
\end{table}


\textbf{OI:} In Table \ref{tab:oi} we show the competition results on both the public and private leader board. The score is computed by weight average of three metrics: recall of top 50 predictions (R@50), mean average precision of relationships (mAP\_rel), mean average precision of phrases (mAP\_phr). The weights for them are 0.2, 0.4, 0.4, respectively. Our model surpasses the 2nd place by 15\% relatively on the public dataset and 20\% relatively on the private dataset.

\textbf{VG:} We present experimental comparison with state-of-the-art methods on Visual Genome dataset in Table \ref{tab:vg}. We use the same train/test splits as in \cite{xu2017scenegraph}. We use the same evaluation  metrics used in \cite{zellers2018neural}, which uses three modes: 1) \textbf{Predicate Classification}: predict predicate labels given ground truth subject and object boxes and labels; 2) \textbf{Scene Graph Classification}: predict subject, object and predicate labels given ground truth subject and object boxes; 3) \textbf{Scene Graph Detection}: predict all the three labels and two boxes. Recalls under the top 20, 50, 100 predictions are used as the measurements.

\textbf{VRD:} We compare with state-of-the-art methods on VRD dataset in Table \ref{tab:vrd}. We use the metrics presented in \cite{yu17iccv}. Note that there is a variable $k$ in this metric which is the number of relation candidates when selecting top50/100. Since not all previous methods specified $k$ in their evaluation, we first report performance in the ``free $k$'' column when considering $k$ as a hyper-parameter that can be cross-validated. For methods where the $k$ is reported for 1 or more values, the column reports the performance using the best $k$. We then list all available results with specific $k$ in the right two columns.

\begin{figure*}[t!]
  \centering
  \begin{subfigure}{\columnwidth}
    \centering
    \begin{subfigure}{0.32\columnwidth}
      \centering
      \adjincludegraphics[width=\textwidth,trim={0 0 0 {.15\height}},clip]{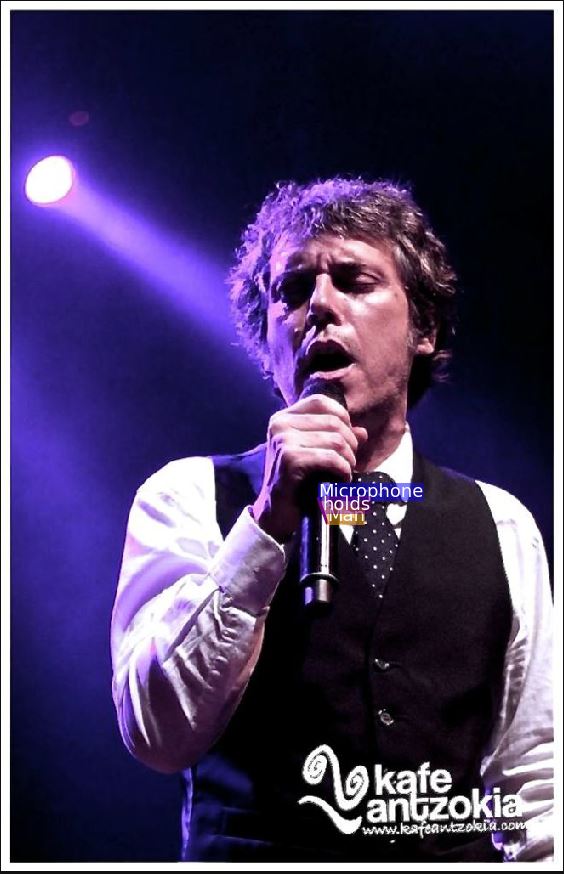}
    \caption{image with gt relationship}
    \label{fig:gt}
    \end{subfigure}
    \centering
    \begin{subfigure}{0.32\columnwidth}
      \centering
      \adjincludegraphics[width=\textwidth,trim={0 0 0 {.15\height}},clip]{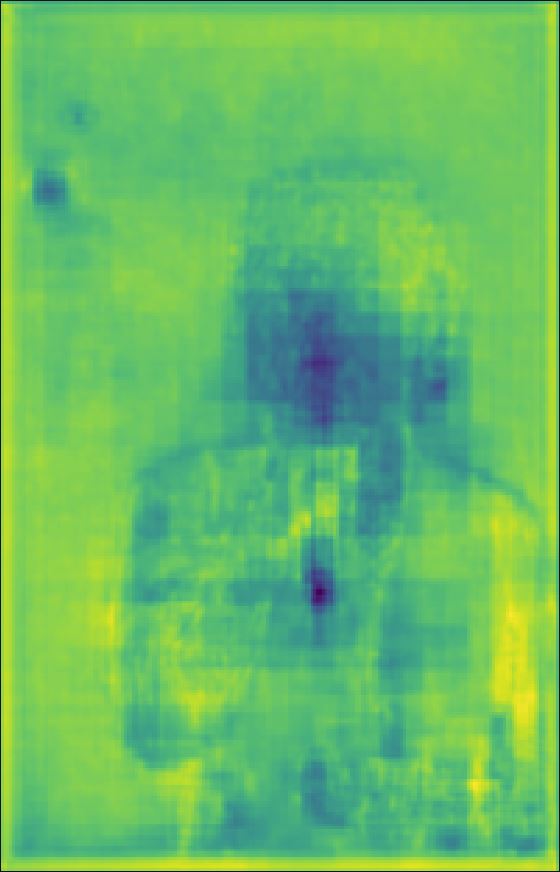}
    \caption{feature from conv\_body\_det}
    \label{fig:det_conv}
    \end{subfigure}
    \centering
    \begin{subfigure}{0.32\columnwidth}
      \centering
      \adjincludegraphics[width=\textwidth,trim={0 0 0 {.15\height}},clip]{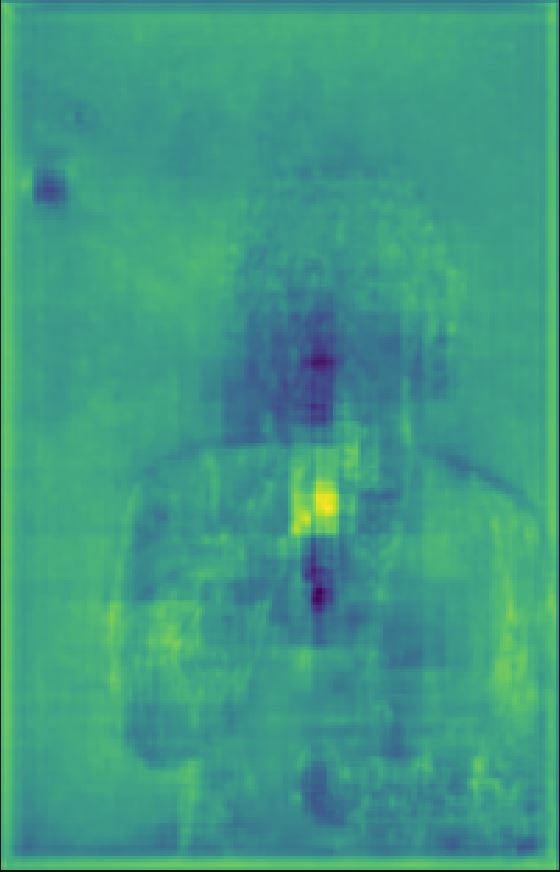}
    \caption{feature from conv\_body\_rel}
    \label{fig:prd_conv}
    \end{subfigure}
  \end{subfigure}
\captionsetup{font=small}
\caption{Visualization of learned CNN features. (a) shows the image with $\langle man,holds,microphone \rangle$, (b) shows the convolution feature from the object detector backbone, and (c) shows the feature from the predicate backbone that we train along with the whole model.}
\label{fig:vis}
\end{figure*}

\textbf{Visualization Results:} In Figure \ref{fig:vis} we show convolution feature maps from the two backbones described in Figure \ref{fig:architecture} given an image with a ground-truth relationship $\langle man,holds,microphone \rangle$. It is very clear that the object detector focuses mostly on the contour of the person, while the predicate branch accurately learns to capture the most informative region that represents ``holds'', i.e., the intersection of the microphone and the fingers that are holding it. This is the most critical reason why our model performs well.

\textbf{Ablation Study:} We show evaluation results on the validation set of four models with the following settings: 1) \textbf{baseline}: only the semantic module. 2) $\langle \textbf{S,P,O} \rangle$: using semantic module and visual module without the direct predictions from subject/object. 3) $\langle \textbf{S,P,O} \rangle \textbf{+ S + O}$: using semantic module and the complete visual module 4) $\langle \textbf{S,P,O} \rangle \textbf{+ S + O + spt}$: our complete model.

\begin{figure*}[t!]
  \centering
  \begin{subfigure}{0.65\columnwidth}
    \centering
    \begin{subfigure}{0.49\columnwidth}
      \centering
      \includegraphics[width=\textwidth]{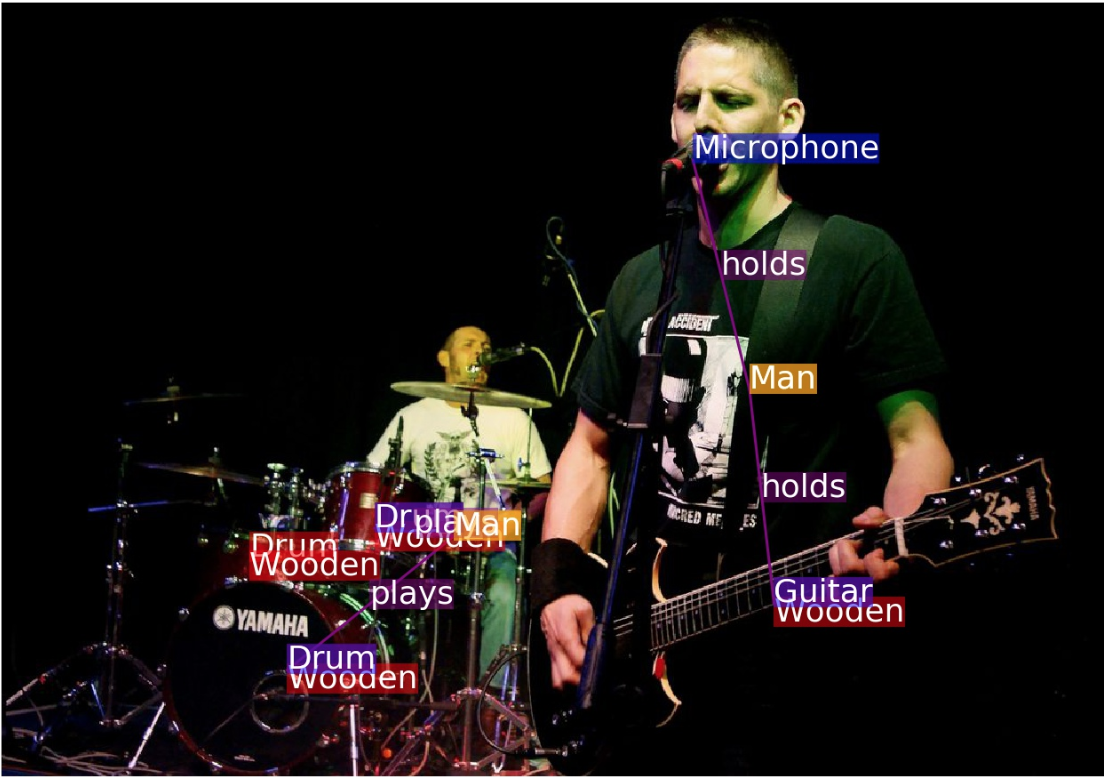}
    \end{subfigure}
    \centering
    \begin{subfigure}{0.49\columnwidth}
      \centering
      \includegraphics[width=\textwidth]{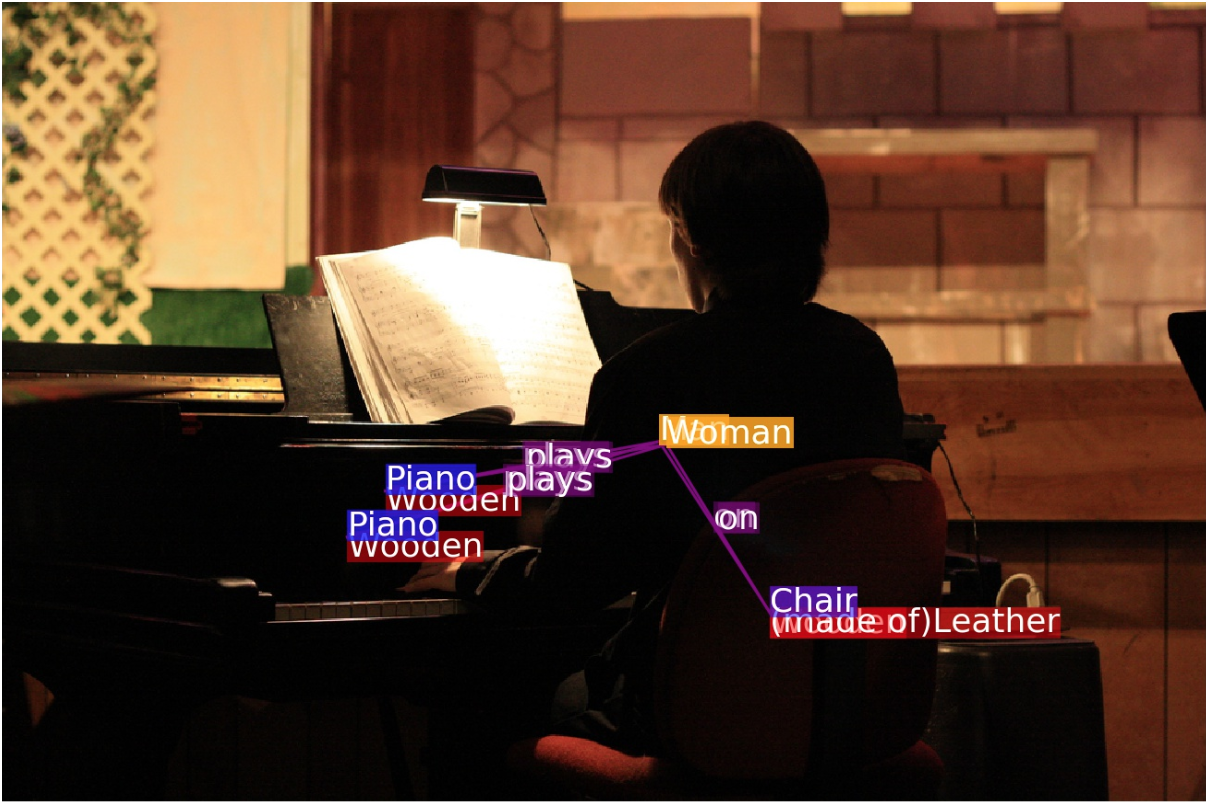}
    \end{subfigure}
    \centering
    \begin{subfigure}{0.49\columnwidth}
      \centering
      \includegraphics[width=\textwidth]{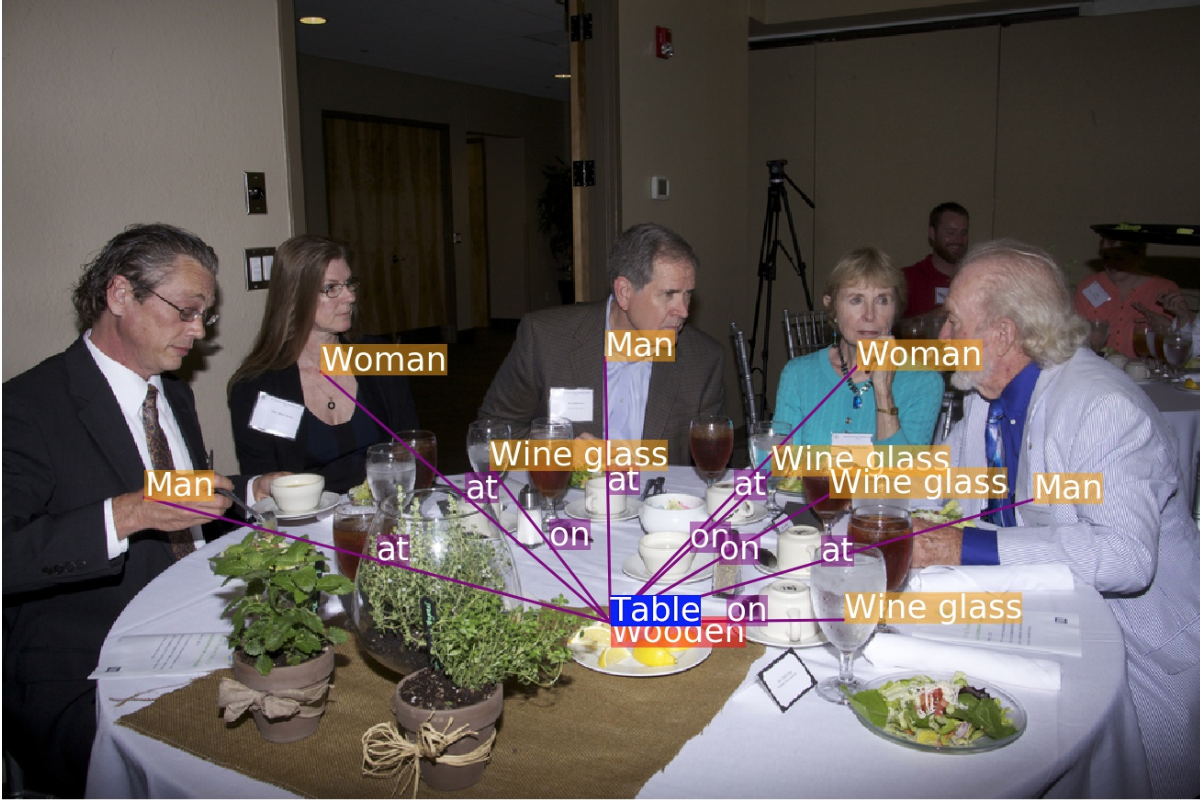}
    \end{subfigure}
    \centering
    \begin{subfigure}{0.49\columnwidth}
      \centering
      \includegraphics[width=\textwidth]{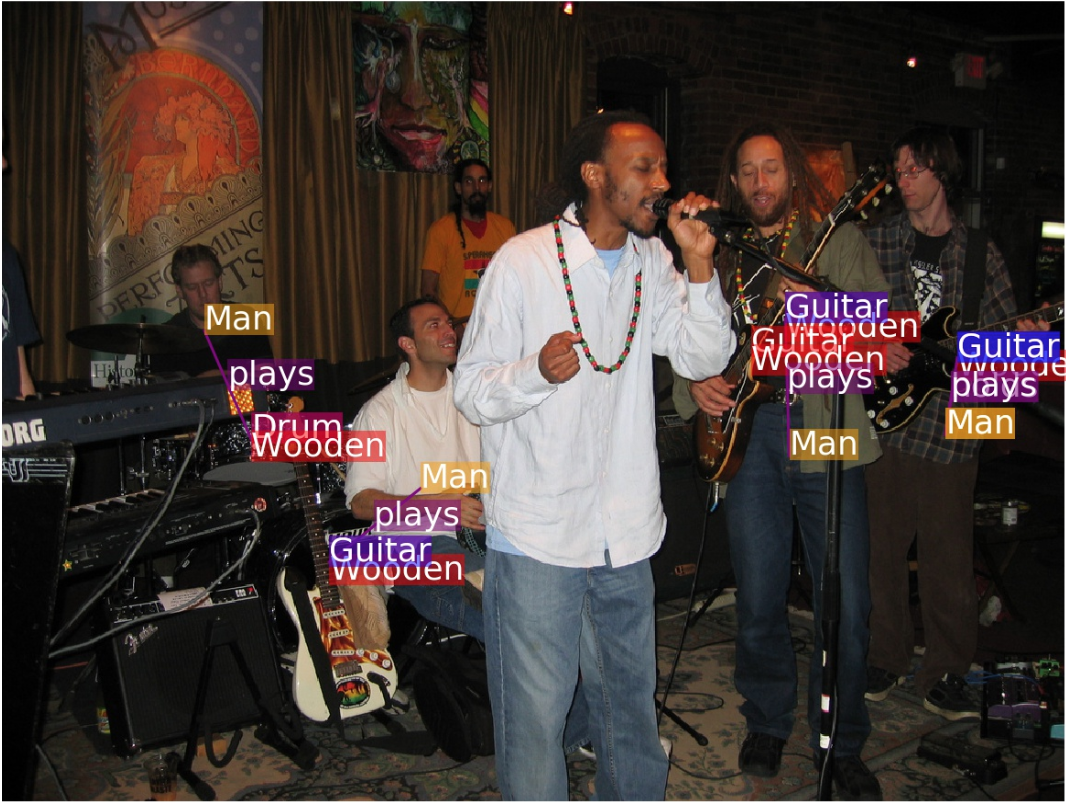}
    \end{subfigure}
  \end{subfigure}
  \centering
  \begin{subfigure}{0.34\columnwidth}
    \centering
    \begin{subfigure}{0.9\columnwidth}
      \centering
      \includegraphics[width=\textwidth]{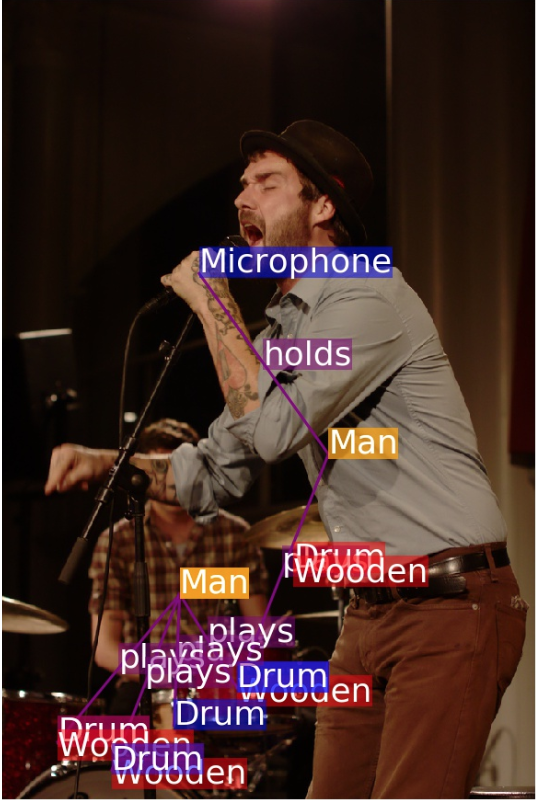}
    \end{subfigure}
  \end{subfigure}
\setlength\belowcaptionskip{-2ex}
\captionsetup{font=small}
\caption{Qualitative results}
\label{fig:qualitative}
\end{figure*}

\textbf{Qualitative Results on OI:} We show several example outputs of our model. We can see from Figure \ref{fig:qualitative} that we are able to correctly refer relationships, i.e., when there are multiple people playing multiple guitars, our model accurately points to the truly related pairs. Our model is also able to handle potentially confusing cases, e.g., in the rightmost image, the person is holding the microphone but not playing the drum, though from a coarse view point it looks like his hand is touching the drum.

{\small
\bibliographystyle{ieee}
\bibliography{egbib}
}

\clearpage

\end{document}